\documentclass[conference]{IEEEtran}
\IEEEoverridecommandlockouts
\usepackage{cite}

\usepackage{amsmath,amssymb,amsfonts}
\usepackage{algorithmic}
\usepackage[ruled]{algorithm2e}
\usepackage{graphicx}
\usepackage{array}
\usepackage{textcomp}
\usepackage{booktabs}
\usepackage[colorlinks=false]{hyperref}
\usepackage{xcolor}
\def\BibTeX{{\rm B\kern-.05em{\sc i\kern-.025em b}\kern-.08em
    T\kern-.1667em\lower.7ex\hbox{E}\kern-.125emX}}
\begin{document}

\title{Label Propagation Through Optimal Transport\\}

\author{\IEEEauthorblockN{Mourad El Hamri}
\IEEEauthorblockA{\textit{LIPN, CNRS UMR 7030} \\
\textit{La Maison des Sciences Numériques}\\
Universit\'e Sorbonne Paris Nord \\
mourad.elhamri@sorbonne-paris-nord.fr}
\and
\IEEEauthorblockN{ Youn\`es Bennani}
\IEEEauthorblockA{\textit{LIPN, CNRS UMR 7030} \\
\textit{La Maison des Sciences Numériques}\\
Universit\'e Sorbonne Paris Nord \\
younes.bennani@sorbonne-paris-nord.fr}
\and
\IEEEauthorblockN{Issam Falih}
\IEEEauthorblockA{\textit{LIMOS, CNRS UMR 6158} \\
\textit{La Maison des Sciences Numériques}\\
Université Clermont Auvergne \\
issam.falih@uca.fr}
}
\maketitle
\begin{abstract}
In this paper, we tackle the transductive semi-supervised learning problem that aims to obtain label predictions for the given unlabeled data points according to Vapnik's principle. Our proposed approach is based on optimal transport, a mathematical theory that has been successfully used  to address various machine learning problems, and is starting to attract renewed interest in semi-supervised learning community. The proposed approach, Optimal Transport Propagation (OTP), performs in an incremental process, label propagation through the edges of a complete bipartite edge-weighted graph, whose affinity matrix is constructed from the optimal transport plan between empirical measures defined on labeled and unlabeled data. OTP ensures a high degree of predictions certitude by controlling the propagation process using a certainty score based on Shannon's entropy. We also provide a convergence analysis of our algorithm. Experiments task show the superiority of the proposed approach over the state-of-the-art. We make our code publicly available.
\footnote{Code is available at: \url{https://github.com/MouradElHamri/OTP}} 
\end{abstract}

\begin{IEEEkeywords}
Optimal Transport, Semi-supervised Learning, Label Propagation
\end{IEEEkeywords}

\section{Introduction}
\setlength{\parindent}{0em}
Deep learning models have achieved state-of-the-art performance on a broad spectrum of learning tasks, and are becoming increasingly popular in various application domains \cite{lecun2015deep}, such
as image classification 
and speech recognition,
where a large amount of labeled data is available. However, for many tasks, it is often prohibitively expensive to collect a large high quality labeled dataset due to lack of time, resources, or other factors, while unlabeled data is cheap and abundant. Medicine is the best illustration of this scenario, where measurement require expensive machinery and labels are the result of a labor and expensive expert-assisted time-consuming analysis. 
\\
\\ In conjunction with transfer learning (TL), semi-supervised learning (SSL) constitute an attractive approach towards addressing the lack of massive labeled datasets. It seeks to largely alleviate the need for labeled samples by providing a means to jointly leverage unlabeled instances. Graph-based semi-supervised approaches are one of the most widely used classes of semi-supervised learning methods, due to their performance and to more and more real graph datasets. The problem is to predict the label of all the unlabeled vertices in the graph based only on a small subset of labeled vertices.
\\A popular graph-based semi-supervised learning method is to use label propagation, this latter has shown good performances in different machine learning applications over the past few years, such as social network analysis \cite{boldi2011layered}  \cite{zhang2017label}, natural language processing \cite{barba2020mulan}, and image segmentation  
\cite{breve2019interactive}.
\\
\\ Most existing label propagation algorithms essentially estimate a data labeling on a fully connected graph constructed by connecting similar samples. The fully connected graph leads typically to a complete labeling for both unlabeled and labeled (re-labeling) samples, which may be interesting under the presence of label noise assumption. Otherwise, it is necessary to add a regularization term to the corresponding objective function in order to penalize predicted labels that do not match the correct ones \cite{van2020survey}. The main existing label propagation approaches can be divided into two categories, methods in the first category such as \cite{zhu2002learning}\cite{zhou2003learning}, capture information on a bilateral level, and methods in the second category as \cite{wang2007label}, capture information on a local level. First category approaches use a Gaussian kernel with a free parameter $\sigma$ to compute the pairwise relationships between data points, which has some drawbacks, since it is hard to determine the optimal value of $\sigma$ if only very few labeled instances are available \cite{zhou2003learning}, and the labeling is very sensitive to the parameter $\sigma$ in the Gaussian kernel \cite{wang2007label}. Instead of the pairwise relationships, which has the disadvantage that only the pairwise relations between the instances are taken into account, second category approaches use the local neighborhood information which assumes that each data point can be optimally reconstructed using a linear combination of its neighbors, which has also many inconveniences, since the optimal number of instances constituting the linear neighborhood must be determined in advance, and even a small variation of its value could make the labeling results very different, without forgetting that the linearity assumption is mainly for computational convenience \cite{wang2007label}. Furthermore, approaches in both categories are unable to capture the underlying geometry of the entire input space and the different interactions that may occur between labeled and unlabeled data points in a global level, and they have another major disadvantage, that of inferring simultaneously all the pseudo-labels by hard assignment, while neglecting the different certainty degree of each prediction. An efficient label propagation approach capable of addressing all these points has not yet been reported.
\\ One of the paradigms used to capture the underlying geometry of the data is grounded on the theory of optimal transport \cite{villani2008optimal} \cite{santambrogio2015optimal}. Optimal  transport provides a powerful means with many attractive theoretical properties for comparing probability measures in a Lagrangian framework, that make many machine learning fields, rely on it to model tasks, compute solutions, and provide theoretical analysis of the algorithms, such as domain adaptation \cite{courty2016optimal}\cite{NIPS2017_0070d23b}\cite{redko2019optimal}\cite{redko2017theoretical}, clustering \cite{laclau2017co}\cite{bouazza2019multi}\cite{bouazza2020collaborative}, generative models \cite{martin2017wasserstein} and more recently semi-supervised learning \cite{solomon2014wasserstein} \cite{taherkhani2020transporting}.
\\
\\ In this paper, we address the existing issues of label propagation methods described above by proposing a principally new approach based on optimal transport that efficiently solves the transductive semi-supervised learning and allows the control of the predictions certainty. We construct an uncommon graph for semi-supervised learning: A complete bipartite edge-weighted graph, to avoid adding a regularization term in the corresponding objective function to penalize predicted labels that do not match the correct ones. In order to benefit from all the geometrical information available in the input space, the affinity matrix of this graph is inferred from the optimal transport plan between empirical measures defined on labeled and unlabeled data points. Furthermore, to take advantage from the reliance of semi-supervised methods to the amount of prior information, we adopt an incremental process to propagate labels through the vertices of the graph, this will allow us to enrich the labeled set with new samples at each iteration, and then, to label the still unlabeled instances with a high certainty. In order to reinforce the certitude of the predictions, we incorporate a certainty score that controls the incremental propagation process. We also provide a convergence analysis for the proposed approach, to show that it requires a finite number of iterations to converge. To the best of our knowledge, the proposed approach is the first optimal transport method for label propagation task.
\\
\\ The rest of this paper is organized as follows: in Section 2, we present an overview of transductive semi-supervised learning. Section 3 details optimal transport problem and its entropic regularized version. In Section 4, we present the proposed OTP method. In Section 5, we provide comparisons to state-of-art methods on seven benchmark datasets. 
\section{Transductive Semi-supervised Learning}
In traditional machine learning, a distinction has usually been made between two major tasks: Supervised and unsupervised learning. Conceptually, semi-supervised learning (SSL) \cite{zhu2005semi} is  situated between them. The aim of semi-supervised learning is to use the abundant amount of unlabeled samples, as well as a typically smaller set of labeled instances, to improve the performance that can be obtained either by discarding the unlabeled data and doing classification (supervised learning) or by discarding the available labels and doing clustering (unsupervised learning). 
\\
\\ Semi-supervised learning makes use of four kind of assumption to utilize the underlying unstructured data. Smoothness assumption: For two data points $x, x^{'}$ that are close in the input space $\mathcal{X}$, the corresponding labels $y, y^{'}$ should be the same. Low-density assumption: The decision boundary should preferably pass through low-density regions in the input space $\mathcal{X}$. Manifold assumption: The high-dimensional input space $\mathcal{X}$ is constituted of multiple lower-dimensional substructures known as manifolds and samples lying on the same manifold should have the same label. Cluster assumption: Data points belonging to the same cluster are likely to have the same label.
\\
\\ According to its objective, semi-supervised learning, can be categorized into two sub-paradigms: Transductive and Inductive semi-supervised learning \cite{van2020survey}. Transductive semi-supervised learning is exclusively interested with obtaining label predictions for the given unlabeled data points. However, inductive semi-supervised learning seeks to  infer a good classifier that can estimate efficiently the label for any instance in the input space, even for previously unseen data points.
\\
\\ In transductive semi-supervised learning settings, we have a finite ordered set of $l$ labeled examples $\{(x_1,y_1),...,(x_l,y_l)\}$. Each example $(x_i,y_i)$ of this set consists of an data point $x_i$ from a given input space $\mathcal{X}$, and its corresponding label $y_i \in \mathcal{Y} = \{c_1,...,c_K\}$, where $\mathcal{Y}$ is a discrete label set composed by $K$ classes. In conjunction with labeled samples, we also have access to a larger collection of $u$ data points $\{x_{l+1},...,x_u\}$, whose labels are unknown. In the remainder, we denote with $X_L$ and $X_U$ respectively the collection of labeled and unlabeled data points, and with $Y_L$ the labels corresponding to $X_L$. Transductive semi-supervised learning aims to infer directly the labels $Y_U$ of the unlabeled instances using all the samples in $X=X_L \cup X_U$ and labels $Y_L$
\cite{chapelle2009semi}. The goal of transductive semi-supervised learning makes it by essence a perfect illustration of Vapnik's principle: When trying to solve some problem, one should not solve a more difficult problem as an intermediate step. Thus instead of inferring a classifier over the input space and evaluating it on the unlabeled points, Vapnik's principle suggests naturally, to propagate information through direct connections between data points, which can be achieved using a graph based method, namely label propagation. 
\\
\\ Label propagation approaches typically involve two phases: A graph construction phase where each instance is represented by a vertex, similar vertices are then connected to each other by edges, thereafter, the edges are weighted to indicate the degree of similarity between the vertices. The edges weights are denoted by an affinity matrix. Second phase is label propagation, where the already constructed graph in the first phase is used to spread labels through its edges, from labeled vertices to unlabeled ones.
\section{Optimal Transport}
In this section we present key concepts of optimal transport and its entropic regularized version. 
\\
\\ Optimal transport \cite{villani2008optimal}\cite{santambrogio2015optimal} is the branch of mathematics that seeks to transform a probability measure into another one while minimizing the total cost of transportation. We owe the first formulation of the optimal transport problem to the French mathematician Gaspard Monge \cite{monge1781memoire}: Let $(\mathcal{X},\mu)$ and $(\mathcal{Y},\nu)$ be two probability spaces, $c$ a positive cost function over $\mathcal{X}\times\mathcal{Y}$, which represents the work needed to transport a unit of mass from $x \in \mathcal{X}$ to $y \in \mathcal{Y}$. The problem asks to find a measurable transport map  $\mathcal{T} : \mathcal{X} \to \mathcal{Y}$ such that:
\begin{equation} (\mathcal{M}) \,\,\,\,\,\,\underset{\mathcal{T}}{\inf}\{\int_{\mathcal{X}}  c(x,\mathcal{T}(x)) d\mu(x) |  \mathcal{T}\#\mu = \nu \},   \end{equation}
where $\mathcal{T}\#\mu$ stands for the image measure of $\mu$ by $\mathcal{T}$. 
The problem $(\mathcal{M})$ is not symmetrical, and may not admit a solution, it is the case when $\mu$ is a Dirac mass and $\nu$ is not.
\\
\\ A convex relaxation of the original problem was suggested by the Soviet mathematician and economist Leonid Kantorovitch \cite{kantorovich1942translocation}, this formulation allows mass splitting and it guarantees to have a solution under very general assumptions.
\begin{equation} (\mathcal{MK}) \,\,\,\,\,\,\underset{\gamma}{\inf} \{\, \int_{\mathcal{X}\times\mathcal{Y}} \, c(x,y) \, d\gamma(x,y) \,|\, \gamma \in \Pi(\mu,\nu)\, \}, \end{equation}
where $\Pi(\mu,\nu)$ is the set of transport plans, constituted of all joint probability measures $\gamma$ on $\mathcal{X}\times\mathcal{Y}$ with marginals $\mu$ and $\nu$: $\Pi(\mu,\nu) = \{ \gamma \in \mathcal{P}(\mathcal{X}\times\mathcal{Y})|\pi_1\#\gamma = \mu$ and $\pi_2\#\gamma = \nu$\}.     
${\pi}_{1}$ and ${\pi}_{2}$ stand for the projection maps:
\begin{center}
$\begin{aligned}[t]
\pi_1 \colon \mathcal{X}\times\mathcal{Y} &\to \mathcal{X} \\
(x,y) &\mapsto x 
\end{aligned}
\quad\text{and}\quad
\begin{aligned}[t]
\pi_2 \colon \mathcal{X}\times\mathcal{Y} &\to \mathcal{Y}. \\
(x,y) &\mapsto y
\end{aligned}$
\end{center}
When $\mathcal{X}=\mathcal{Y}$ is a metric space equipped with a distance $d$, it is natural to use it as a cost function, e.g. $c(x, y) = d(x, y)^p$ for $p \in {\left[1\,,+\infty\right[}$. In such case, the problem $(\mathcal{MK})$ defines a metric between probability measures over $\mathcal{X}$, called the $p$-Wasserstein distance, defined as follows,  $\forall \mu,\nu \in \mathcal{P}(\mathcal{X})$:
\begin{equation}
    \mathrm{W}_{p}(\mu,\nu) = \underset{\gamma \in \Pi(\mu,\nu)}{\inf} (\int_{\mathcal{X}^{2}} d^{p}(x,y) \, d\gamma(x,y))^{1/p}, 
\end{equation}
Wasserstein distance has an intuitive formulation as well as the ability to capture the underlying geometry of the measures by relying on the metric $d$. Wasserstein distance metrize weak convergence, and it allows comparison between probability measures, even when the supports of the measures do not overlap. These attractive properties make it an ideal candidate for learning problems.
\\
\\ In the discrete version of the optimal transport problem, i.e. when $\mu$ and $\nu$ are only available through discrete samples $X = (x_1,...,x_n) \subset \mathcal{X}$ and $Y = (y_1,...,y_m) \subset \mathcal{Y}$, the empirical distributions can be taken to be discrete measure: $\mu = \sum_{i=1}^n a_{i} \delta_{x_{i}}$ and
$\nu = \sum_{j=1}^m b_{j} \delta_{y_{j}}$, where  $a=(a_1,...,a_n)$ and $b=(b_1,...,b_m)$ are vectors in the probability simplex $\sum_n$ and $\sum_m$ respectively, where: $\sum_k = \{ u \in \mathbb{R}^k \, | \, \forall i \le k, \, u_i \ge 0 \, \,\text{and} \, \sum_{i=1}^k   u_i =1 \}$. The cost function only needs to be specified for every
pair $(x_i,y_j) \in X \times Y$ yielding a cost matrix $C \in  \mathcal{M}_{n \times m}(\mathbb{R}^{+})$. The optimal transport problem becomes then a linear program \cite{bertsimas1997introduction}, parametrized by the cost matrix $C$ and the transportation polytope $U(a,b) = \{\gamma \in \mathcal{M}_{n \times m}(\mathbb{R}^{+}) \, | \, \gamma 1_{m} = a \,\, \text{and} \,\, \gamma^{\mathbf{T}} 1_{n} = b\}$, which acts as a feasible set. Thus, solving this linear program consists of finding a plan $\gamma^*$ that realizes:
\begin{equation}
(\mathcal{D}_\mathcal{MK}) \,\,\,\,\,\,\underset{\gamma \in U(a,b)}{\min}  \langle {\gamma},{C} \rangle _F,
\end{equation}
where $\langle.,.\rangle_F$ is the Frobenius dot product.
\\ 
\\ If $n = m$, and, $\mu$ and $\nu$ are uniform measures, $U(\mathbf{1}_n/n,\mathbf{1}_n/n)$ is then the Birkhoff polytope of size $n$, and the solutions of $(\mathcal{D}_\mathcal{MK})$ are permutation matrices.
\\
\\ Discrete optimal transport is a linear program, and thus can be solved exactly in $\mathcal{O}(n^3 log(n))$ when comparing two discrete measures of $n$ points with interior point methods \cite{pele2009fast}, which is a heavy computational price tag. Alternatively, one can consider an entropic regularized version of the problem, which allows a very fast computation of the transport plan.
\\
\\ In \cite{cuturi2013sinkhorn}, Cuturi proposed to add an entropic regularization
term to the expression of the discrete optimal transport problem. The regularized version of the discrete optimal transport reads:
\begin{equation}
    \underset{\gamma \in U(a,b)}{\min}  \langle {\gamma},{C} \rangle _F - \varepsilon \mathcal{H}(\gamma)   ,
\end{equation}
where  $\mathcal{H}(\gamma) = - \sum_{i=1}^n \sum_{j=1}^m \gamma_{ij} (\log(\gamma_{ij}) - 1)  $ is the entropy of $\gamma$.
\\
\\ Since the function $\mathcal{H}$ is 1-strongly concave: Its Hessian is $\partial^2 \mathcal{H}(\gamma) = -diag(\frac{1}{\gamma_{ij}})$ \text{and} $\gamma_{ij} \le 1.$ Then, the objective of the regularized optimal transport is an $\varepsilon$-strongly convex function, thus the regularized problem has a unique optimal solution.
The solution of the regularized optimal transport problem has the form ${\gamma_\varepsilon^*} = diag(u)Kdiag(v)$, where $u$ and $v$ are the exponential scaling of the dual variables corresponding to each marginal constraint, and $K$ is the exponential scaling of the cost matrix $C$. The variables $(u,v)$ must therefore satisfy: $u\odot (Kv) = a  \, \, \, and  \, \, \, v\odot(K^{\mathbf{T}}u) = b$, where $\odot$ corresponds to entrywise multiplication of vectors. This problem is known in the numerical analysis community as the matrix scaling problem \cite{nemirovski1999complexity}, and can be solved efficiently via an iterative procedure: The Sinkhorn-Knopp algorithm \cite{knight2008sinkhorn}, which iteratively update $ u^{(l+1)} = \frac{a}{Kv^{(l)}} , \, \, \, \text{and}  \, \, \,    v^{(l+1)} = \frac{b}{K^{\mathbf{T}}u^{(l+1)}},$ initialized with an arbitrary positive vector $v^{(0)}= 1_{m}$.
\\
\\ For a small regularization $\varepsilon$, the unique solution $\gamma_\varepsilon^*$ of the regularized problem converges with respect to the weak topology to the optimal solution with maximal entropy within the set of all optimal solutions of $(\mathcal{D}_\mathcal{MK})$ \cite{peyre2019computational}.
\section{OPTIMAL TRANSPORT PROPAGATION : OTP}
In this section, we introduce the proposed OTP
approach. The main underlying idea behind OTP is to use the optimal transport plan between the empirical measures defined on labeled and unlabeled instances in order to construct an improved affinity matrix, and then, to use it to propagate labels from labeled to unlabeled data points in an incremental process ensuring prediction certainty.
\subsection{Problem setup}
Let $X=\{x_1,...,x_{l+u}\}$ be a set of $l+u$ data points in  the input space $\mathbb{R}^d$ and $\mathcal{C} =\{c_1,...,c_K\}$ a discrete label set consisting of $K$ classes. The first $l$ points denoted by $X_L=\{x_1,...,x_l\}$ are labeled according to $Y_L=\{y_1,...,y_l\}$, where $y_i \in \mathcal{C}$ for every $i \in \{1,...,l\}$, and the remaining samples denoted by $X_U=\{x_{l+1},...,x_{l+u}\}$ are unlabeled. Usually $l \ll u$. The aim of label propagation algorithms is to infer the unknown labels $Y_U$ using all the data points in $X=X_L \cup X_U$ and labels $Y_L$.
\\
\\ To use an appropriate formulation to the paradigm of optimal transport, the empirical distribution of $X_L$ and $X_U$ must be expressed respectively using discrete measures as:
\begin{equation}
\mu = \sum_{i=1}^l a_{i} \delta_{x_{i}} \, \, \,\, \text{and} \, \, \,\, \nu = \sum_{j=l+1}^{l+u} b_{j} \delta_{x_{j}},
\end{equation}
Under the assumption that $X_L$ and $X_U$ are a collection of independent and identically distributed data points, the weights of all instances in each sample are naturally set to be equal:
\begin{equation} a_i = \frac{1}{l},\forall i \in \{1,...,l\} \, \text{and} \, b_j = \frac{1}{u},\forall j \in \{l+1,...,l+u\},
\end{equation}
\subsection{Proposed approach}
Label propagation algorithms are often  graph-based methods, typically consisting of two phases, the first one is the graph construction and the second is the label propagation phase, where we spread labels from the labeled vertices of the graph already constructed in the first phase, to unlabeled ones.
\\
\\ 
\textbf{1\textsuperscript{st} phase:} The main underlying idea of the first phase is the use of a complete bipartite edge-weighted graph \cite{asratian1998bipartite}  $\mathcal{G}=(\mathcal{V},\mathcal{E},\mathcal{W})$, where $\mathcal{V} = X$ is the vertex set, that can be divided into two disjoint and independent sets $\mathcal{L} = X_L$ and $\mathcal{U} = X_U$, $\mathcal{E} \subset \{\mathcal{L} \times \mathcal{U}\}$ is the edge set, and $\mathcal{W} \in \mathcal{M}_{l,u}(\mathbb{R}^+)$ is the affinity matrix to denote the edges weights. The weight $w_{i,j}$ on edge $e_{i,j} \in \mathcal{E}$ reflects the degree of similarity between $x_i \in  \mathcal{L}$ and $x_j \in \mathcal{U}$. We propose to infer the affinity matrix $\mathcal{W}$ from the optimal transport plan between the measures $\mu$ and $\nu$. The intuition behind the use of this uncommon type of graph in semi-supervised learning, is to exploit its ability to assign labels only for instances in $\mathcal{U}$, without modifying the labels of samples in $\mathcal{L}$, which makes it very attractive for label propagation tasks outside the label noise assumption, i.e. there is no need to add any regularization term to the objective function of the label propagation algorithm in order to penalize predicted labels that do not match the correct ones in $\mathcal{L}$.
\\
\\ To measure quantitatively the similarity between vertices, we need to use some distance over the input space $\mathbb{R}^d$. Let $C \in \mathcal{M}_{l,u}(\mathbb{R}^+)$ denotes the matrix of squared euclidean distances between vertices in $\mathcal{L}$ and $\mathcal{U}$, defined as follows:
\begin{equation}
c_{i,j} = \lVert x_i - x_j \rVert^2,  \, \, \, \forall (x_i,x_j ) \in \mathcal{L} \times \mathcal{U},
\end{equation}
In order to construct an affinity matrix $\mathcal{W}$ that captures the underlying geometry of the whole data $X$ in the input space and all the interactions between labeled and unlabeled data in a global vision, instead of the pairwise relationships that perform in a bilateral level or the local neighborhood information and to avoid the use of a Gaussian kernel, a natural choice is to rely on the optimal transport theory. Since optimal transport suffers from a computational burden, we can overcome this issue by using its entropic regularized version, in the following way:
\begin{equation}
\gamma^*_\varepsilon = \underset{\gamma \in U(a,b)}{argmin} \,\, \langle {\gamma},{C} \rangle _F - \varepsilon \mathcal{H}(\gamma)   ,
\end{equation}
    The optimal transport plan $\gamma^*_\varepsilon$ provides us the weights of associations between vertices in $\mathcal{L}$ and $\mathcal{U}$, thus, $\gamma^*_\varepsilon$ can be interpreted in our context as a similarity matrix between the two parts $\mathcal{L}$ and $\mathcal{U}$ of the graph $\mathcal{G}$: Similar labeled and unlabeled vertices correspond to higher value in $\gamma^*_\varepsilon$.
\\
\\ To have a class probability interpretation, we column-normalize the matrix $\gamma^*_\varepsilon$, it will give a non-square left-stochastic affinity matrix $\mathcal{W}$, defined as follows:
\begin{equation}
w_{i,j} = \frac{\gamma^*_{\varepsilon_{i,j}}}{\sum_i \gamma^*_{\varepsilon_{i,j}}}, \, \, \, \forall i,j  \in \{1, ..., l\}\times\{l+1, ..., l+u \},
\end{equation}
where $w_{i,j},\, \, \,  \forall i,j  \in \{1, ..., l\} \times \{l+1, ..., l+u \}$ is then, the probability of jumping from the vertex $x_i \in \mathcal{L}$ to $x_j\in \mathcal{U}$.
\\
\\
\textbf{2\textsuperscript{nd} phase:} Our intuition is to use the affinity matrix  $\mathcal{W}$ in the second phase to identify labeled data points who should spread their labels to similar unlabeled instances. We suggest to use an incremental process to label the points in $\mathcal{U}$. We suggest also to provide with each pseudo-label a certainty score that measures the certitude of the prediction, and to use it in order to control the incremental label propagation process.
\\
\\ First, we need to construct a label matrix $U \in \mathcal{M}_{u,K}(\mathbb{R}^+)$, to denotes the probability of each unlabeled data point $x_j$, $j \in \{l+1, ..., l+u \}$ to belong to a class $c_k$, $k \in \{1, ..., K\}$. For a harmonious construction of the label matrix $U$ with the information coming from the optimal transport plan $\gamma^*_\varepsilon$, we propose to define the probability of an unlabeled data point $x_j$ to belong to a class $c_k$ as the sum of its similarity with the representatives of this class: 
\begin{equation}
u_{j,k} = \sum_{i / x_i \in c_k} w_{i,j} ,  \forall j,k \in \{l+1, ..., l+u \} \times \{1, ..., K\},
\end{equation}
The matrix $U$ is a non-square right-stochastic matrix, and can be interpreted as a vector-valued function $U : X_U \rightarrow \sum_K$, which assigns a stochastic vector $U_j \in \sum_K$ to each unlabeled data point $x_j$, $j \in \{l+1, ..., l+u \}$. 
\\
\\ Traditional label propagation approaches infer simultaneously all the pseudo-labels by hard assignment, without worrying about the fact that these label predictions do not have the same degree of certainty. This issue, as mentioned by \cite{iscen2019label}, can degrade significantly the performance of the label propagation approaches. To prevent this, we suggest to associate a certainty score $s_j$ with the label prediction of each $x_j, j \in \{l+1, ..., l+u \}$. The proposed certainty score $s_j$ is defined in the following way:
\begin{equation}
s_j = 1 -  \frac{H(Z_j)}{log_2(K)}, \, \, \, \, \, \forall j \in \{l+1, ..., l+u \},
\end{equation}
where $Z_j : \mathcal{C} \rightarrow  \mathbb{R}$ is a real-valued random variable, defined by $Z_j(c_k) =k$, to associate a numerical value $k$ to the potential label prediction result $c_k$. The probability distribution of the random variable $Z_j$ is encoded in the stochastic vector $U_j$:
\begin{center}
    $\mathbb{P}(Z_j = c_k) =u_{j,k}, \,\, \, \, \, \, \forall j,k \in \{l+1, ..., l+u \} \times \{1, ..., K\}$
\end{center}
$H$ is an uncertainty measure, which we suggest to take equal to Shannon's entropy \cite{shannon2001mathematical}, defined by: 
\begin{center}
    $H(Z_j)  = - \sum_k u_{j,k}\log_2(u_{j,k})$,
\end{center}
we divide $H$ by $log_2(K)$ to normalize it between $0$ and $1$.
\\
\\ To control the certainty of the propagation process, we define a confidence threshold $\alpha \in [0, 1]$, and for each unlabeled data point $x_j$, we make a comparison between $\alpha$ and $s_j$. If the score $s_j$ is greater than $\alpha$, we assign to $x_j$ a pseudo-label $\hat{y_{j}}$, in the following way:
\begin{equation}
\hat{y_{j}} =  \underset{c_{k} \in \mathcal{C}}{argmax} \, u_{j,k}, \, \,  \forall j \in \{l+1, ..., l+u \},
\end{equation}
Thus, the unlabeled instance $x_j$ will belong to the class  $c_k$ with the highest class-probability $u_{j,k}$, in other words, to the class whose representatives possess the highest similarity with it. Otherwise, we do not give any label to the point $x_j$.
\\
\\ The process above corresponds to one iteration of the proposed incremental approach. At each of its iterations, $X_L$ is enriched with new instances, and the number of data points in $X_U$ is reduced. This modification of $X_L$, $Y_L$ and $X_U$ resulting from the incremental approach is of major importance in the context of label propagation, since, the effectiveness of a label propagation algorithm depends on the amount of prior information, thus,  increasing the size of $X_L$ at each iteration, will similarly increase the performance of the proposed approach, and will make it possible to label the points still in $X_U$ with a high degree of certainty at the next iterations. We repeat the same whole procedure at each iteration until convergence, here convergence means that all the data initially in $X_U$ are labeled during this incremental process. The proposed algorithm, named OTP, is formally summarized in Algorithm $1$:
\begin{algorithm}\label{OTP}
\caption{OTP}
\SetKwInOut{Input}{Input}
\SetKwInOut{Parameters}{Parameters}
\SetKwInOut{Output}{Output}
\Parameters{$\varepsilon, \alpha $}
\Input{$X_L, X_U, Y_L$}
\While{not converged}{
Compute the cost matrix $C$ by Eq(8) \\
Solve the optimal transport problem in Eq(9) \\
Compute the affinity matrix $\mathcal{W}$ by Eq(10)\\
Get the label matrix $U$ by Eq(11) \\
\For{$x_j \in X_U$}{
Compute the certainty score $s_j$ by Eq(12) \\
\eIf{$s_j > \alpha $}{
    Get the pseudo label $\hat{y_{j}}$ by Eq(13)\\
    Inject $x_j$ in $X_L$ \\
    Inject $\hat{y_{j}}$ in $Y_L$ \\
    
}{
Maintain $x_j$ in $X_U$ \\
}
}
}
\Return{$Y_U$}
\end{algorithm}
\begin{figure*}[h]
	\centering
\includegraphics[width=1.\textwidth]{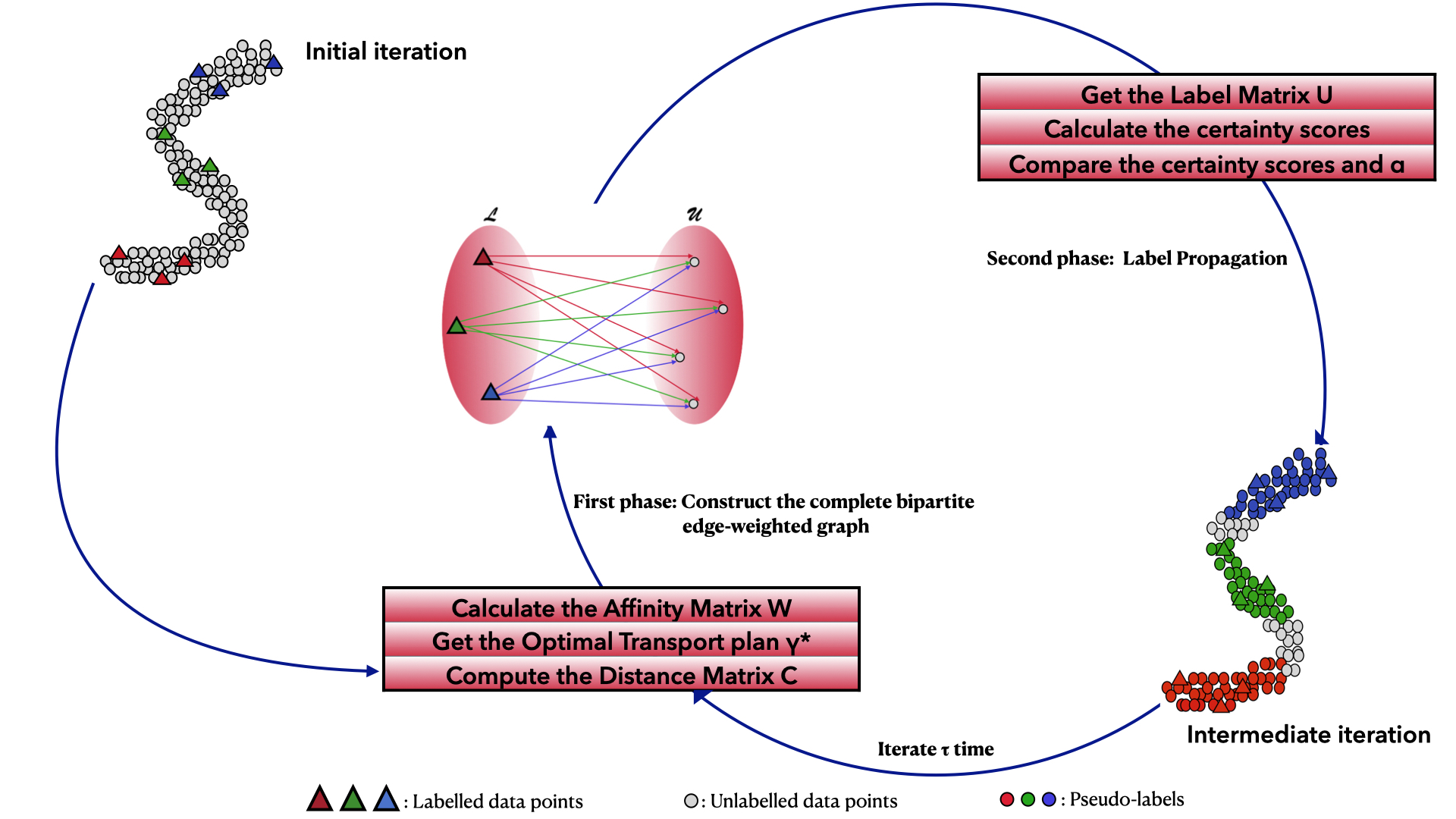}
	\label{fig:pccccs}
	\caption{Overview of OTP. We initiate an incremental approach where at each iteration, we construct a complete bipartite edge-weighted graph based on the optimal transport plan between the distribution of labeled instances and unlabeled ones. Then, we propagate labels through the edges of the graph. Triangles markers correspond to the labeled instances and circles correspond to the unlabeled data which are gradually pseudo-labeled by OTP. The class is color-coded.}
\end{figure*}
\subsection{Convergence analysis }
As mentioned earlier, the convergence of OTP means that all the samples initially in $X_U$ are labeled during the incremental process. In order to analyze the convergence of the proposed approach, we suggest to express the evolution of  $X_L$ and $X_U$ respectively as follows: Let $m_t$ be the size of $X_L$ and $n_t$ be the size of $X_U$ at an iteration $t$, then the evolution of the two sets can be expressed using the following dynamical systems:
\begin{equation} 
\begin{cases} 
m_t  = m_{t-1} + \zeta_t\\
m_0 = l 
\end{cases},  
\quad 
\begin{cases} 
n_t = n_{t-1} - \zeta_t\\
n_0 = u 
\end{cases}   
\end{equation}
where $\zeta_t$ is the number of instances in $X_U$ that have been labeled during the iteration $t$.
\\
\\ Theoretically, OTP must converge at the iteration $\tau$, where : $m_{\tau} = m_0 + \sum_{t=1}^{\tau} \zeta_t = m_0 + u =l + u$, which corresponds also to $n_{\tau} = n_0 - \sum_{t=1}^{\tau}  \zeta_t = n_0 - u = u -u= 0$.
\\
\\The convergence analysis of our approach corresponds to showing that the iteration $\tau$ will be reached in a finite number of intermediate iterations. To achieve this objective, a suitable choice of $\alpha$ must be made to allow us to label a large amount $\zeta_t$ of instances in $X_U$ at each iteration $t$. However, the following scenario can be encountered in some intermediate iterations: Suppose that at an iteration $t$, we still have $h$ unlabeled instances in $X_U$, whose certainty score $s_j$ is lower than the threshold $\alpha$,
which means that none of these instances can be labeled at this iteration. To overcome this issue, one can think to decrease the value of the threshold $\alpha$ to allow only to the point with the greatest certainty score among the $h$ unlabeled instances, to be labeled, and then to move from $X_U$ to $X_L$. This can be achieved as follows:
\begin{equation}
    \alpha \leftarrow \alpha - \min_{x_j \in [X_U]_t}  (\alpha - s_j),
\end{equation}
where $[X_U]_t$ is the set of the $h$ points constituting $X_U$ at the intermediate iteration $t$. Certainly, decreasing sharply the value of $\alpha$ will allow us to label many other instances instead of just the data point with the highest certainty score, however, this gain in terms of the number of points labeled at the same iteration will be paid out in terms of its predictions certainty. Our intuition behind the modification of $\alpha$ in the way above is as follows: Since moving an instance from $X_L$ to $X_U$, can radically change the optimal transport plan between the new distributions $\mu$ and $\nu$, and subsequently the certainty scores in the next iteration,  we can try to restore the initial value of $\alpha$ and continue to label the other samples with the same degree of certainty as before. If the same scenario is repeated in a future iteration, we can use the same technique of decreasing $\alpha$ to label a new point, and so on until convergence. This reasoning shows that the proposed algorithm needs effectively a finite number of iterations to converge.
\section{EXPERIMENTAL RESULTS}
In this section, we provide empirical experimentation for the proposed algorithm. 
\subsection{Datasets} 
The experiment was designed to evaluate the proposed approach on seven benchmark datasets. Details of these  datasets appear in the following table:
\begin{table}[h]
\caption{Experimental datasets}
\centering
\setlength\extrarowheight{-1.pt}
\begin{tabular}{lccc}
\hline
Datasets  \quad \quad   & \#Instances  \quad \quad  & \#Features \quad \quad& \#Classes  \\  
\toprule 
Iris        & 150       & 4      & 3          \\ 
Heart       & 270       & 13     & 2          \\ 
Ionosphere  & 351       & 34     & 2          \\ 
Dermatology & 366       & 33     & 6          \\
Waveform    & 5000      & 21     & 3          \\ 
Digits      & 5620      & 64     & 10         \\ 
MNIST       & 10000     & 784    & 10         \\
\toprule
\end{tabular}
\label{Data}
\end{table}
\subsection{Evaluation indices}
In order to evaluate the performance of our approach, two evaluation measures were employed: The normalized mutual information \cite{dom2012information}, and the adjusted rand index \cite{hubert1985comparing}. These two evaluation indices seek to measure the similarity between two partitions on a dataset. Given a dataset $D$ with $N$ instances and two partitions of these samples, namely the ground-truth partition $P= \{p_1, p_2,..., p_K\}$ and the  partition resulting from the label propagation $P^{'}=\{p_1^{'},p_2^{'},...,p_K^{'}\}$. Let $n_{ij} = |p_i^{'} \cap p_j|$ be the number of common nodes of groups $p_i^{'}$ and $p_j$, $b_i = \sum_{j=1}^N n_{ij}$ and $d_j = \sum_{i=1}^N n_{ij}$.
\\
\\ The normalized mutual information is defined as :
\begin{center}
    NMI = $\frac{2 \sum_i \sum_j n_{ij} \log \frac{n_{ij}N}{b_id_j} }{-\sum_i b_i\log \frac{b_i}{N} -\sum_j  d_j \log \frac{d_j}{N}}$
\end{center}
The adjusted rand index is defined as :
\begin{center}
    ARI = $\frac{\sum_{ij} \binom{n_{ij}}{2} - [\sum_i \binom{b_{i}}{2}\sum_j \binom{d_{j}}{2}]/\binom{N}{2}}{\frac{1}{2}[\sum_i \binom{b_{i}}{2} + \sum_j \binom{d_{j}}{2}]-[\sum_i \binom{b_{i}}{2}\sum_j \binom{d_{j}}{2}]/\binom{N}{2}}$
\end{center}
If the label propagation partition is close to the true partition, then its NMI and ARI values are close to 1.
\subsection{Experimental protocol}
The proposed algorithm were compared with three label propagation approaches, including LP \cite{zhou2003learning} and LS \cite{zhu2002learning}, which are the classical label propagation algorithms, LNP \cite{wang2007label}, which is another label propagation algorithm with an improved affinity matrix, and with CNMF\cite{liu2010non} which is an NMF based constrained clustering method and PLCC \cite{liu2017partition} which is a $k$-means based partition level constrained clustering method. To compare the seven approaches, their related parameters were specified as follows: Each of the compared algorithms LP, LS and NLP, require a Gaussian function with a free parameter $\sigma$ to be determined in order to built their affinity matrix. 
In the comparisons, each of these three algorithms was tested with different values of $\sigma$, and its best value corresponding to the highest NMI and ARI values on each dataset was selected. The number of clusters $k$ was set equal to the true number of classes on each dataset for CNMF and PLCC. The performance of a label propagation approach depends on the available amount of prior information. Thus, in the experiment, the amount of prior information data was set to 5, 15, 25, and 35 percent of the total number of samples in the datasets. The performance of a label propagation approach depends also on the quality of prior information. Therefore, in the experiment, given the amount of prior information, all the six compared algorithms were run with $10$ different sets of prior information to compute the average results for NMI and ARI on each dataset. The comparison present also the average performances of each approach over all the datasets.
\subsection{Experimental results}
The following tables list the performance of the six approaches on all the datasets. Experiments confirm that the prior information is able to improve the labeling effectiveness, in fact, given a dataset, all the label propagation and the partition level constrained clustering algorithms show a growth in their performance with respect to both NMI and ARI, in parallel with the increase of the amount of prior information. Furthermore, the tables show that the proposed approach is clearly more accurate than LP, LS, NLP, CNMF, and PLCC on all the tested datasets. The tables also present the average results of each algorithm, which confirm that the proposed label propagation approach based on optimal transport outperforms the other methods on all the datasets, followed by LS, LP, NLP, PLCC and then CNMF, in that order.
\\
\\ This results are mainly attributed to the capacity of OTP to capture mush more information than the other methods thanks to the enhanced affinity matrix constructed by optimal transport. It is also worth noting that, the performance of OTP lies in the fact that the incremental process take advantage of 
\newpage
\begin{table}[h]
\caption{NMI values for Transductive Semi-supervised methods}
\small
\centering
\setlength\tabcolsep{1.3pt}
\setlength\extrarowheight{-2pt}
\begin{tabular}{lccccccc}
\hline
Datasets    & Percent  & LP      & LS      & LNP     & CNMF    & PLCC     & OTP            \\ \toprule 
            & 5\%      & 0.7302  & 0.7354  & 0.6729  &0.4561   &0.5187    & \textbf{0.7372} \\ 
            & 15\%     & 0.8412  & 0.8442  & 0.7534  &0.5274   &0.5835    & \textbf{0.8447} \\ 
Iris        & 25\%     & 0.8584  & 0.8621  & 0.8269  &0.5717   &0.6489    & \textbf{0.8667}  \\             
            & 35\%     & 0.8621  & 0.8649  & 0.8314  &0.6198   &0.7067    & \textbf{0.8852} \\

            & 5\%   & 0.0991 & 0.1037 & 0.0721        &0.0439    &0.0828    &\textbf{0.1394} \\
            & 15\%  & 0.1519 & 0.1575 & 0.1091        &0.1163    &0.1390    &\textbf{0.2181} \\
Heart       & 25\%  & 0.2291 & 0.2472 & 0.1432        &0.1858    &0.2021    &\textbf{0.3683} \\
            & 35\%  & 0.3313 & 0.3546 & 0.2718        &0.3087    &0.3191    &\textbf{0.4374} \\

            & 5\%   & 0.2837  & 0.2895 & 0.2510     &0.1329    &0.2157   & \textbf{0.3786} \\
            & 15\%  & 0.3502  & 0.3535 & 0.3256     &0.2278    &0.3007   & \textbf{0.4676} \\ 
Ionosphere  & 25\%  & 0.3848  & 0.3911 & 0.3572     &0.2605    &0.3356   & \textbf{0.5000} \\ 
            & 35\%  & 0.3972  & 0.4014 & 0.3725     &0.2892    &0.3529   & \textbf{0.5383}\\ 
        
            & 5\%   & 0.7927  & 0.8054 & 0.7785     &0.5197    &0.6665   & \textbf{0.8159} \\
            & 15\%  & 0.8770  & 0.8779 & 0.8349     &0.5531    &0.6991   & \textbf{0.8935} \\
Dermatology & 25\%  & 0.8932  & 0.8932 & 0.8692     &0.6238    &0.7201   & \textbf{0.9033} \\ 
            & 35\%  & 0.9128  & 0.9128 & 0.8959     &0.6703    &0.7732   & \textbf{0.9164} \\  

           & 5\%   & 0.4102  & 0.4184 & 0.3981      &0.1829    &0.2761  & \textbf{0.4493}   \\
           & 15\%  & 0.4950  & 0.5009 & 0.4628      &0.2453    &0.3191  & \textbf{0.5256}  \\ 
Waveform   & 25\%  & 0.5124  & 0.5192 & 0.4763      &0.2619    &0.3307  & \textbf{0.5319} \\ 
           & 35\%  & 0.5192  & 0.5229 & 0.4807      &0.2792    &0.3391  & \textbf{0.5421} \\ 
          
           & 5\%   & 0.8319  & 0.8319 & 0.7850      &0.1471    &0.6798  & \textbf{0.8571}   \\
           & 15\%  & 0.9150  & 0.9150 & 0.8891      &0.1617    &0.7412  & \textbf{0.9290} \\ 
Digits     & 25\%  & 0.9443  & 0.9443 & 0.9268      &0.2435    &0.7801  & \textbf{0.9489} \\ 
           & 35\%  & 0.9570  & 0.9570 & 0.9318      &0.3174    &0.7956  & \textbf{0.9607} \\ 

           & 5\%  & 0.7429 & 0.7436 & 0.7005        &0.1037    &0.5616  & \textbf{0.7581}\\
           & 15\% & 0.8019 & 0.8028 & 0.7759        &0.2452    &0.6329  & \textbf{0.8177} \\
MNIST      & 25\% & 0.8389 & 0.8367 & 0.7931        &0.2912    &0.6506  & \textbf{0.8442}  \\
           & 35\% & 0.8542 & 0.8599 & 0.8136        &0.3201    &0.6711  & \textbf{0.8730}\\
        \toprule 
ALL Datasets & Average  & 0.6363  & 0.6409 & 0.5999 &0.3180    &0.5015      & \textbf{0.6767} \\ \toprule 
\end{tabular}
\label{NMI}
\vspace{-5mm}
\end{table} 

\begin{table}[h]
\caption{ARI values for Transductive Semi-supervised methods}
\small
\centering
\setlength\tabcolsep{1.3pt}
\setlength\extrarowheight{-2pt}
\begin{tabular}{lccccccc}
\hline
Datasets & Percent   & LP      & LS     & LNP   & CNMF & PLCC &  OTP  \\ \toprule

             & 5\%   & 0.7481  & 0.7428 &0.7062  &0.4165 &0.4725 &\textbf{0.7582} \\ 
             & 15\%  & 0.8453  & 0.8492 &0.7861  &0.4986 &0.5403 &\textbf{0.8621} \\ 
Iris         & 25\%  & 0.8680  & 0.8704 &0.8321  &0.5215 &0.6029 &\textbf{0.8884} \\ 
             & 35\%  & 0.8754  & 0.8783 &0.8424  &0.5791 &0.6471 &\textbf{0.9027} \\

            & 5\%   & 0.1385 & 0.1120 & 0.0892        &0.0938    &0.1108    &\textbf{0.1694} \\
            & 15\%  & 0.2110 & 0.2190 & 0.1562        &0.1492    &0.1979    &\textbf{0.2475} \\
Heart       & 25\%  & 0.3176 & 0.2955 & 0.2283        &0.2201    &0.2593    &\textbf{0.4162} \\
            & 35\%  & 0.4163 & 0.4464 & 0.3688        &0.3350    &0.3728    &\textbf{0.5030} \\ 
             
             & 5\%   & 0.3081  & 0.3124 &0.2828  &0.1625 &0.2752 &\textbf{0.5089}\\
             & 15\%  & 0.4221  & 0.4248 &0.3998  &0.2541 &0.3491 &\textbf{0.5723} \\ 
Ionosphere   & 25\%  & 0.4606  & 0.4673 &0.4324  &0.3005 &0.3491 &\textbf{0.5927}\\ 
             & 35\%  & 0.4650  & 0.4702 &0.4418  &0.3217 &0.3902 &\textbf{0.6281} \\ 
             
             & 5\%   & 0.7834  & 0.8121 & 0.7808 &0.5360 &0.6826 &\textbf{0.8289} \\ 
             & 15\%  & 0.8807  & 0.8813 & 0.8438 &0.5725 &0.7174 &\textbf{0.8996} \\ 
Dermatology  & 25\%  & 0.8972  & 0.8972 & 0.8751 &0.6401 &0.7486 &\textbf{0.9093}  \\                         
             & 35\%  & 0.9146  & 0.9146 & 0.9007 &0.6935 &0.7910 &\textbf{0.9218} \\  
             
             & 5\%   & 0.4728  & 0.4771 &0.4494&0.2201 &0.3025 &\textbf{0.5084} \\
             & 15\%  & 0.5639  & 0.5678 &0.5163&0.2819 &0.3486 &\textbf{0.5945} \\ 
Waveform     & 25\%  & 0.5819  & 0.5864 &0.5279&0.3059 &0.3618 &\textbf{0.6031} \\ 
             & 35\%  & 0.5870  & 0.5880 &0.5342&0.3242 &0.3745 &\textbf{0.6182} \\ 
             
             & 5\%   & 0.8290  & 0.8290 &0.7964&0.0982 &0.6169 &\textbf{0.8684} \\
             & 15\%  & 0.9126  & 0.9127 &0.8993&0.1174 &0.6931 & \textbf{0.9306} \\ 
Digits       & 25\%  & 0.9432  & 0.9432 &0.9287&0.1834 &0.7306 &\textbf{0.9508}  \\ 
             & 35\%  & 0.9567  & 0.9567 &0.9407&0.2587 &0.7294 &\textbf{0.9621} \\ 
             
             & 5\%   & 0.7510  & 0.7523 &0.7106&0.0539 &0.4518 &\textbf{0.7763} \\
             & 15\% & 0.7930 & 0.7944 &0.7697&0.0970 &0.5692 &\textbf{0.8393} \\
MNIST        & 25\% &  0.8487 & 0.8466 &0.8152&0.1193 &0.5927 &\textbf{0.8685} \\
             & 35\% & 0.8721 & 0.8777 &0.8438&0.1452 &0.6201 &\textbf{0.8935} \\
             
\toprule 
ALL Datasets  & Average  & 0.6665  & 0.6687 & 0.5467 &0.2674 &0.4963 & \textbf{0.7151} \\ \toprule 
\end{tabular}
\label{ARI}
\vspace{-18mm}
\end{table} 
\newpage

the dependency of label propagation algorithms on the amount of prior information, then the enrichment of the labeled set at each iteration with new instances, allows to the unlabeled samples to be labeled with a high degree of certainty at next iterations. We can also explain the improvement provided by our approach to its ability to control the certitude of the label predictions thanks to the certainty score used, which allows to data points to be labeled only if they have a high degree of prediction certainty. 
\section{CONCLUSION}
In this paper we proposed OTP, a novel method dealing with transductive semi-supervised learning notably label propagation problem. Our method is principally different from other label propagation methods and consists in inferring an improved affinity matrix from the optimal transport plan between labeled and unlabeled instances. An incremental procedure was used to take advantage of the dependency of label propagation methods to the amount of prior information, and a certainty score was incorporated to assure the certainty of predictions during the label propagation process. Experiments have shown that OTP approach outperforms current state-of-the-art methods. In the future, we plan to extend OTP to inductive settings and to use the pseudo-labels inferred by OTP with the initial labeled data to train a CNN model in a deep semi-supervised  manner for computer vision tasks.
\bibliographystyle{plain}
\bibliography{IJCNN2021.bib}
\end{document}